%% file: root.tex
\documentclass[letterpaper, 10 pt, conference]{ieeeconf}  

\IEEEoverridecommandlockouts                              

\usepackage{cite}
\usepackage{amsmath}
\usepackage{amssymb}
\usepackage{amsfonts}
\usepackage{algorithmic}
\usepackage[ruled,linesnumbered,vlined]{algorithm2e}
\usepackage[final]{graphicx}
\usepackage{textcomp}
\usepackage{xcolor}
\usepackage{etoolbox}
\usepackage{subcaption}
\usepackage{comment}
\usepackage{cleveref}

\usepackage{mathtools}
\title{\LARGE \bf
Learning a Control Policy for Fall Prevention on an Assistive Walking Device
}

\author{
Visak CV Kumar$^{1}$, Sehoon Ha$^{2}$, Gregory Sawicki$^{1}$, C. Karen Liu$^{3}$
\thanks{$^{1}$ Georgia Institute of Technology, Atlanta, GA, 30308, USA}
\thanks{$^{2}$ Robotics at Google, Mountain View, CA, 94043, USA }
\thanks{$^{3}$ Stanford University, Stanford, CA, 94305, USA}
\thanks{Email: {\tt\small visak3@gatech.edu, sehoonha@google.com gregory.sawicki@me.gatech.edu, karenliu@cs.stanford.edu}}
}

\begin{document}

\maketitle
\thispagestyle{empty}
\pagestyle{empty}

\begin{abstract}
Fall prevention is one of the most important components in senior care. We present a technique to augment an assistive walking device with the ability to prevent falls. Given an existing walking device, our method develops a fall predictor and a recovery policy by utilizing the onboard sensors and actuators. The key component of our method is a robust human walking policy that models realistic human gait under a moderate level of perturbations. We use this human walking policy to provide training data for the fall predictor, as well as to teach the recovery policy how to best modify the person's gait when a fall is imminent. Our evaluation shows that the human walking policy generates walking sequences similar to those reported in biomechanics literature. Our experiments in simulation show that the augmented assistive device can indeed help recover balance from a variety of external perturbations. We also provide a quantitative method to evaluate the design choices for an assistive device.      
\end{abstract}

   
\input{defs.tex}
\input{Introduction.tex}
\input{Related.tex}
\input{Method.tex}

\input{Experiment.tex}

\input{Conclusion.tex}

\bibliographystyle{IEEEtran}
\bibliography{annot}

\end{document}

%% file: defs.tex

\newcommand{\cmt}[1]{}
\newcommand{\visak}[1]{\textcolor{blue}{{Visak: #1}}}
\newcommand{\sehoon}[1]{\textcolor{magenta}{{Sehoon: #1}}}
\newcommand{\karen}[1]{\textcolor{red}{{Karen: #1}}}
\newcommand{\newtext}[1]{#1}
\newcommand{\original}[1]{\textcolor{magenta}{Original: #1}}
\newcommand{\eqnref}[1]{Equation~(\ref{eq:#1})}
\newcommand{\figref}[1]{Figure~\ref{fig:#1}}
\newcommand{\algref}[1]{Algorithm~\ref{alg:#1}}
\newcommand{\tabref}[1]{Table~\ref{tab:#1}}
\newcommand{\secref}[1]{Section~\ref{sec:#1}}

\long\def\ignorethis#1{}

\newcommand{\etal}{{\em{et~al.}\ }}
\newcommand{\eg}{e.g.\ }
\newcommand{\ie}{i.e.\ }

\newcommand{\figtodo}[1]{\framebox[0.8\columnwidth]{\rule{0pt}{1in}#1}}



\newcommand{\pdd}[3]{\ensuremath{\frac{\partial^2{#1}}{\partial{#2}\,\partial{#3}}}}

\newcommand{\mat}[1]{\ensuremath{\mathbf{#1}}}
\newcommand{\set}[1]{\ensuremath{\mathcal{#1}}}

\newcommand{\vc}[1]{\ensuremath{\mathbf{#1}}}
\newcommand{\vEndEff}{\ensuremath{\vc{d}}}
\newcommand{\vRelMove}{\ensuremath{\vc{r}}}
\newcommand{\sSet}{\ensuremath{S}}

\newcommand{\vControl}{\ensuremath{\vc{u}}}
\newcommand{\vPoint}{\ensuremath{\vc{p}}}
\newcommand{\sSpringCoef}{{\ensuremath{k_{s}}}}
\newcommand{\sDamperCoef}{{\ensuremath{k_{d}}}}
\newcommand{\vHandle}{\ensuremath{\vc{h}}}
\newcommand{\vForce}{\ensuremath{\vc{f}}}

\newcommand{\mTransChain}{\ensuremath{\vc{W}}}
\newcommand{\mRotateTrans}{\ensuremath{\vc{R}}}
\newcommand{\sJoint}{\ensuremath{q}}
\newcommand{\vJoint}{\ensuremath{\vc{q}}}
\newcommand{\mJoint}{\ensuremath{\vc{Q}}}
\newcommand{\mMass}{\ensuremath{\vc{M}}}
\newcommand{\sMass}{\ensuremath{{m}}}
\newcommand{\vGravity}{\ensuremath{\vc{g}}}
\newcommand{\vConstr}{\ensuremath{\vc{C}}}
\newcommand{\sConstr}{\ensuremath{C}}
\newcommand{\vCOM}{\ensuremath{\vc{x}}}
\newcommand{\sGeneralForce}[1]{\ensuremath{Q_{#1}}}
\newcommand{\vStateVar}{\ensuremath{\vc{y}}}
\newcommand{\vControlVar}{\ensuremath{\vc{u}}}
\newcommand{\tr}[1]{\ensuremath{\mathrm{tr}\left(#1\right)}}

%
%

\renewcommand{\choose}[2]{\ensuremath{\left(\begin{array}{c} #1 \\ #2 \end{array} \right )}}

\newcommand{\gauss}[3]{\ensuremath{\mathcal{N}(#1 | #2 ; #3)}}

\newcommand{\pctab}{\hspace{0.2in}}
\newenvironment{pseudocode} {\begin{center} \begin{minipage}{\textwidth}
                             \normalsize \vspace{-2\baselineskip} \begin{tabbing}
                             \pctab \= \pctab \= \pctab \= \pctab \=
                             \pctab \= \pctab \= \pctab \= \pctab \= \\}
                            {\end{tabbing} \vspace{-2\baselineskip}
                             \end{minipage} \end{center}}
\newenvironment{items}      {\begin{list}{$\bullet$}
                              {\setlength{\partopsep}{\parskip}
                                \setlength{\parsep}{\parskip}
                                \setlength{\topsep}{0pt}
                                \setlength{\itemsep}{0pt}
                                \settowidth{\labelwidth}{$\bullet$}
                                \setlength{\labelsep}{1ex}
                                \setlength{\leftmargin}{\labelwidth}
                                \addtolength{\leftmargin}{\labelsep}
                                }
                              }
                            {\end{list}}
\newcommand{\newfun}[3]{\noindent\vspace{0pt}\fbox{\begin{minipage}{3.3truein}\vspace{#1}~ {#3}~\vspace{12pt}\end{minipage}}\vspace{#2}}

\newcommand{\key}{\textbf}
\newcommand{\fun}{\textsc}



%% file: Introduction.tex
\section{INTRODUCTION}

More than three million older adults every year in the United States are treated for fall injuries. In 2015, the medical costs for falls amounted to more than \$50 billion. Compounding to the direct injuries, fall-related accidents have long-lasting impact because falling once doubles one's chances of falling again. Even with successful recovery, many older adults develop fear of falling, which may make them reduce their everyday activities. When a person is less active, their health condition plummets which increases their chances of falling again.

Robotic assistive walking devices or exoskeletons are designed to improve the user's ability to ambulate \cite{rubenstein}. Previous work has shown that these devices can increase the gait stability and efficiency when worn by older adults or people with disabilities \cite{Thatte2019,Wu2015,Thatte2015,GALLE2017183}. In this work, we explore the possibility to augment the existing assistive walking devices with the capability to prevent falls while respecting the functional and ergonomical constraints of the device.

Designing a control policy to prevent falls on an existing wearable robotic device has multiple challenges. First, the control policy needs to run in real-time with limited sensing and actuation capabilities dictated by the walking device. Second, a large dataset of human falling motions is difficult to acquire and unavailable to public to date, which imposes fundamental obstacles to learning-based approaches. Lastly and perhaps most importantly, the development and evaluation of the fall-prevention policy depends on intimate interaction with human users. The challenge of modeling realistic human behaviors in simulation is daunting, but the risk of testing on real humans is even greater.

We tackle these issues by taking the approach of model-free reinforcement learning (RL) in simulation to train a fall-prevention policy that operates on the walking device in real-time, as well as to model the human locomotion under disturbances. The model-free RL is particularly appealing for learning a fall-prevention policy because the problem involves non-differentiable dynamics and lacks existing examples to imitate. In addition, demonstrated by recent work in learning policies for human motor skills \cite{peng2018deepmimic,YuSIGGRAPH2018}, the model-free RL provides a simple and automatic approach to solving under-actuated control problems with contacts, as is the case of human locomotion. To ensure the validity of these models, we compare the key characteristics of human gait under disturbances to those reported in the biomechanics literature \cite{Winter,Wang2014}.

Specifically, we propose a framework to automate the process of developing a \emph{fall predictor} and a \emph{recovery policy} on an assistive walking device, by only utilizing the onboard sensors and actuators. When the fall predictor predicts that a fall is imminent based on the current state of the user, the recovery policy will be activated to prevent the fall and deactivated when the stable gait cycle is recovered. The core component of this work is a robust \emph{human walking policy} a moderate level of perturbations. We use this human walking policy to provide training data for the fall predictor, as well as to teach the recovery policy how to best modify the person's gait to prevent falling.

Our evaluation shows that the human policy generates walking sequences similar to the real-world human walking data both with and without perturbation. We also show quantitative evaluation on the stability of the recovery policy against various perturbation. In addition, our method provides a quantitative way to evaluate the design choices of assistive walking device. We analyze and compare the performance of six different configurations of sensors and actuators, enabling the engineers to make informed design decisions which account for the control capability prior to manufacturing process.

%% file: Related.tex
\section{Related Work}

\label{sec:rw}

\subsection{Control of assistive devices}
Many researchers have developed control algorithms for robotic assistive walking devices. 
As reported by Yan \etal~\cite{YAN2015120}, existing methods can be broadly classified into trajectory tracking controllers \cite{Wangpre2015,JezPre2004,BlayaPre} and model-based controllers \cite{WickModel,KazModel,vanderkooijModel}.
Although trajectory tracking approaches can be easily applied to regular walking cycles, it is unclear how to generate trajectories for unexpected situations due to perturbations.
On the other hand, model-based controllers can capture adaptive behaviors by developing control laws with respect to external perturbations but they often require an accurate dynamic model.
Based on neuromuscular model of human lower limbs, Thatte et al \cite{Thatte2015,Thatte2017} presented an optimization-based framework which  converges better than other control strategies, such as quasi-stiffness control \cite{lenzi}, minimum jerk swing control \cite{lenziminjerk}, virtual constraint control \cite{Greggvirtual} and impedance control \cite{supimpedence}. However, all of these methods focused on controlling regular walking cycles rather than adjusting to external perturbations.
More recently, Thatte \etal \cite{Thatte2019} presented a real-time trip avoidance algorithm which estimates future knee and hip positions using a Gaussian filter and updates a trajectory using a fast quadratic program solver. 

\subsection{Deep RL for assistive devices}
Many researchers have demonstrated learning effective policies for high-dimensional control problems using deep reinforcement learning (deep RL) techniques  \cite{schulman2015trust,schulman2017proximal}.
However, there has been limited work at the intersection of deep RL and control of assistive devices. Hamaya \etal \cite{HAMAYA201767} presented model-based reinforcement learning algorithm to train a policy for a handheld device that takes muscular effort measured by electromyography signals (EMGs) as inputs. The work requires collecting user interaction data on the real-world device to build a model of the user's EMG patterns. However, collecting a large amount of data for lower-limb assistive devices is less practical due to the safety concerns.
Another recent work, Bingjing \etal \cite{Bingjing2019}, developed an adaptive-admittance model-based control algorithm for a hip-knee assistive device, the reinforcement learning aspect of this work focused on learning parameters of the admittance model. 
Our method is agnostic to the assistive walking devices and can be used to augment any device that allows for feedback control.

\subsection{Simulating human behaviors}
Our recovery policy assumes that the person is able to walk under a moderate level of perturbation, similar to the real healthy people or people assisted by a walking device. The early work of Shiratori \etal \cite{Shiratori2009} developed a trip recovery controller during walking from motion capture data. However, it requires careful analysis of human responses to design state machines.
Recently, deep RL techniques have been proven to be effective for developing control policies that can reproduce natural human behaviors in simulated environments.
The proposed techniques vary from formulating a reward function from motion capture data \cite{peng2018deepmimic}, to training a low-energy policy with curriculum learning \cite{YuSIGGRAPH2018}, to developing a full-body controller for a muscular-skeleton model \cite{Lee:2019:SMH:3306346.3322972}, to modeling biologically realistic torque limits \cite{Jiang:2019:SBR:3306346.3322966}. 
Particularly, our work is inspired by the DeepMimic technique proposed by Peng \etal \cite{peng2018deepmimic} which produces a visually pleasing walking gait by mimicking the reference motion and shows stable control near the target trajectory.

\subsection{Human responses to perturbations} 
Our framework relies on accurate modelling of human motions, particularly in response to external perturbations.
Researchers in biomechanics studied postural responses of human bodies to falling and slipping incidents and identified underlying principles \cite{Lockhart2012,Moyer2009}.
Connor \etal \cite{connor09} and Hof \etal \cite{Hof} also studied balancing strategies of humans for rejecting medial-lateral perturbations during walking. 
Particularly, we validate the learned human policy by comparing its footstep placement against the data collected by Wang \etal \cite{Wang2014}, which reports a strong correlation between the pelvic states and the next footstep locations.

%% file: Method.tex
\section{Method}
\label{sec:method}
We propose a framework to automate the process of augmenting an assistive walking device with the capability of fall prevention. Our method is built on three components: a human walking policy, a fall predictor, and a recovery policy. We formulate the problem of learning human walking and recovery policies as Markov Decision Processes (MDPs), $(\mathcal{S}, \mathcal{A}, \mathcal{T}, r, p_0, \gamma)$, where $\mathcal{S}$ is the state space, $\mathcal{A}$ is the action space, $\mathcal{T}$ is the transition function, $r$ is the reward function, $p_0$ is the initial state distribution and $\gamma$ is a discount factor. We take the approach of model-free reinforcement learning to find a policy $\pi$, such that it maximizes the accumulated reward:
\begin{equation}
    J(\pi) = \mathbb{E}_{\mathbf{s}_0, \mathbf{a}_0, \dots, \mathbf{s}_T} \sum_{t=0}^{T} \gamma^t r(\mathbf{s}_t, \mathbf{a}_t),\nonumber
\end{equation}
 where $\mathbf{s}_0 \sim p_0$, $\mathbf{a}_t \sim \pi(\mathbf{s}_t)$ and $\mathbf{s}_{t+1}=\mathcal{T}(\mathbf{s}_t, \mathbf{a}_t)$.

We denote the human walking policy as $\pi_{h}(\vc{a}_{h}|\vc{s}_{h})$ and the recovery policy as $\pi_{e}(\vc{a}_{e}|\vc{s}_{e})$, where $\vc{s}_h$, $\vc{a}_h$, $\vc{s}_e$, and $\vc{a}_e$,  represent the corresponding states and actions, respectively. Our method can be applied to assitive walking devices with any sensors or actuators, though we assume that the observable state $\vc{s}_{e}$ of the walking device is a subset of the full human state $\vc{s}_{h}$ due to sensor limitations. Since our method is intended to augment an assistive walking device, we also assume that the user who wears the device is capable of walking. Under such an assumption, our method only needs to model normal human gait instead of various pathological gaits.

\begin{figure}
\centering
\setlength{\tabcolsep}{1pt}
\renewcommand{\arraystretch}{0.7}
\begin{tabular}{c c}
 
  \includegraphics[width=0.19\textwidth,height=3cm]{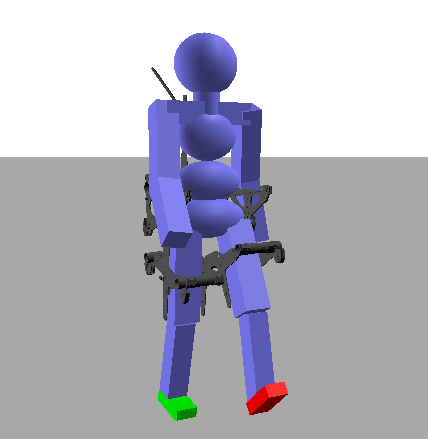}&
  \includegraphics[width=0.19\textwidth,height=3cm]{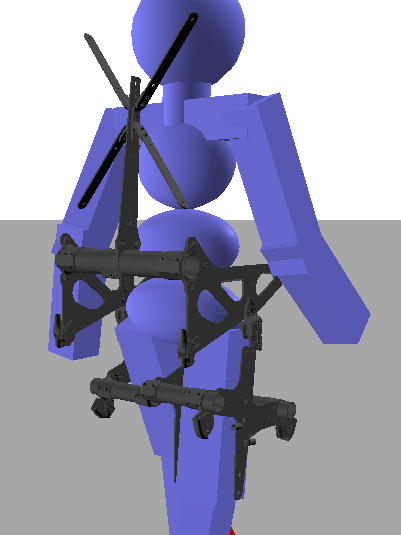} \\ 
  
\end{tabular}
\caption{\textbf{Left :} We model a 29-Degree of Freedom(DoF) humanoid and the 2-DoF exoskeleton in PyDart. \textbf{Right :} Assistive device design used in our experiments. }
\label{fig:Dart}
\end{figure}






\subsection{Human Walking Policy}
\label{sec:human_policy}
We take the model-free reinforcement learning approach to developing a human locomotion policy $\pi_{h}(\vc{a}_{h}|\vc{s}_{h})$. To achieve natural walking behaviors, we train a policy that imitates the human walking reference motion similar to Peng \etal \cite{peng2018deepmimic}. The human 3D model (agent) consists of $23$ actuated joints with a floating base as shown in \figref{Dart}. This gives rise to a $53$ dimensional state space $\vc{s}_{h} = [\vc{q},\vc{\dot{q}},\vc{v}_{com}, \boldsymbol{\omega}_{com},\phi]$, including joint positions, joint velocities, linear and angular velocities of the center of mass (COM), and a phase variable that indicates the target frame in the motion clip. We model the intrinsic sensing delay of a human musculoskeletal system by adding a latency of $40$ milliseconds to the state vector before it is fed into the policy. The action determines the target joint angles $\vc{q}_t^{target}$ of the proportional-derivative (PD) controllers, deviating from the joint angles in the reference motion: 
\begin{equation}
    \label{eq:action}
    \vc{q}^{target}_{t} = \hat{\vc{q}}_{t}(\phi) + \vc{a}_{t},
\end{equation}
where $\hat{\vc{q}}_{t}(\phi)$ is the corresponding joint position in the reference motion at the given phase $\phi$. Our reward function is designed to imitate the reference motion:
\begin{multline}
    \label{eqn:reward}
    r_{h}(\vc{s}_{h},\vc{a}_{h}) =
    w_{q}(\vc{q} - \hat{\vc{q}}(\phi)) \\ 
    + w_{c}(\vc{c} - \hat{\vc{c}}(\phi)) + w_{e}(\vc{e} - \hat{\vc{e}}(\phi)) - w_{\tau}||\boldsymbol{\tau}||^{2},
\end{multline}
where $\hat{\vc{q}}$, $\hat{\vc{c}}$, and $\hat{\vc{e}}$ are the desired joint positions, COM positions, and end-effector positions from the reference motion data, respectively. The reward function also penalizes the magnitude of torque $\boldsymbol{\tau}$. We use the same weight $w_{q} = 5.0$, $w_{c} = 2.0$, $w_{e} = 0.5$, and $w_{\tau} = 0.005$ for all experiments. We also use early termination of the rollouts, if the agent's pelvis drops below a certain height, we end the rollout and re-initialize the state.

Although the above formulation can produce control policies that reject small disturbances near the target trajectory, they often fail to recover from perturbations with larger magnitude, such as those encountered during locomotion. It is critical to ensure that our human walking policy can withstand the same level of perturbation as a capable real person, so that we can establish a fair baseline to measure the increased stability due to our recovery policy. 

Therefore, we exert random forces to the agent during policy training. Each random force has a magnitude uniformly sampled from $[0,800]\ N$ and a direction uniformly sampled from [-$\pi/2$,$\pi/2$], applied for $50$ milliseconds on the agent's pelvis in parallel to the ground. The maximum force magnitude induces a velocity change of roughly $0.5$m/sec. We also randomize the time when the force is applied within a gait cycle. Training in such a stochastic environment is crucial for reproducing the human ability to recover from a larger disturbance during locomotion. We represent a human policy as a multi-layered perceptron (MLP) neural network with two hidden layers of $128$ neurons each. The formulated MDP is trained with Proximal Policy Optimization (PPO) [PPO].

\subsection{Fall Predictor}

Being able to predict a fall before it happens gives the recovery policy critical time to alter the outcome in the near future. We take a data-drive approach to train a classifier capable of predicting the probability of the fall in the next $40$ milliseconds. Collecting the training data from the real world is challenging because induced human falls can be unrealistic and dangerous/tedious to instrument. As such, we propose to train such a classifier using only simulated human motion. Our key idea is to automatically label the future outcome of a state by leveraging the trained human policy $\pi_h$. We randomly sample a set of states $\vc{s}_h$ and add random perturbations to them. By following the policy $\pi_h$ from each of the sampled states, we simulate a rollout to determine whether the state leads to successful recovery or falling. We then label the corresponding state observed by the walking device, $(\vc{s}_e, 1)$ if succeeds, or $(\vc{s}_e, 0)$ if fails. We collect about $50000$ training samples. Note that the input of the training data corresponds to the state of the walking device, not the full state of human, as the classifier will only have access to the information available to the onboard sensors.

We train a support vector machine (SVM) classifier with radial basis function kernel to predict if a state $\vc{s}_{e}$ has a high chance of leading to a fall or not. We perform a six-fold validation test on the dataset and the classifier achieves an accuracy above $94$\%.

\subsection{Recovery Policy}
The recovery policy aims to utilize the onboard actuators of the assistive walking device to stabilize the gait such that the agent can continue to walk uninterruptedly. The recovery policy $\pi_e$ is trained to provide optimal assistance to the human walking policy when a fall is detected. The state of $\pi_e$ is defined as $\vc{s}_{e}=[\dot{\boldsymbol{\omega}},\boldsymbol{\omega},\vc{q}_{hip},\dot{\vc{q}}_{hip}]$, which comprises of global angular acceleration, angular velocity, and hip joint angle position and velocity, amounting to $10$ dimensional state space. The action space consists of torques at two hip joints. The reward function maximizes the quality of the gait while minimizing the impact of disturbance:
\begin{equation}
    \label{eq:exoReward}
    r_{e}(\vc{s}_h,\vc{a}_e) = r_{walk}(\vc{s}_h) - w_{1}\|\vc{v}_{com}\| - w_{2}\|\boldsymbol{\omega}_{com}\| - w_3 \|\vc{a}_e\|,
\end{equation}
where $r_{walk}$ evaluates walking performance using Equation \ref{eqn:reward} except for the last term, and $\vc{v}_{com}$ and $\boldsymbol{\omega}_{com}$ are the global linear and angular velocities. Note that the input to the reward function includes the full human state $\vc{s}_h$. While the input to the recovery policy $\pi_e$ should be restricted by the onboard sensing capability of the assistive walking device, the input to the reward function can take advantage of the full state of the simulated world, since the reward function is only needed at training time. The policy is represented as a MLP neural network with two hidden layers of 64 neurons each and trained with PPO.

%% file: Experiment.tex
\section{Experiments and Results} \label{sec:Exp}

We validate the proposed framework using the open-source physics engine DART \cite{lee2018dart}. Our human agent is modeled as an articulated rigid body system with $29$ degrees of freedom (dofs) including the six dofs for the floating base. The body segments and the mass distribution are determined based on a $50$th percentile adult male in North America. We select the prototype of our assistive walking device as the testbed.Similar prototypes are described in \cite{caputo,ExoDesign,witte}. It has two cable-driven actuators at hip joints, which can exert about $200$ Nm at maximum. However, we limit the torque capacity to $30$ Nm as a hard constraint. Sensors, such as Inertial Measurement Units (IMU) and hip joint motor encoders, are added to the device. We also introduce a sensing delay of $40$ to $50$ ms. We modeled the interaction between the device and human by adding positional constraints on the thigh and anchor points.  For all experiments, the simulation time step is set to $0.002$s. 

We design experiments to systematically validate the learned human behaviors and effectiveness of the recovery policy. Particularly, our goal is to answer the following questions: 
\begin{enumerate}
    \item How does the motion generated by the learned human policy compare to data in the biomechanics literature? 
    \item Does the recovery policy improve the robustness of the gaits to external pushes?
    \item How does the effectiveness of the assistive walking device change with design choices?
\end{enumerate}

\subsection{Comparison of Policy and Human Recovery Behaviors}

\begin{figure}
\centering
\includegraphics[width=0.8\linewidth]{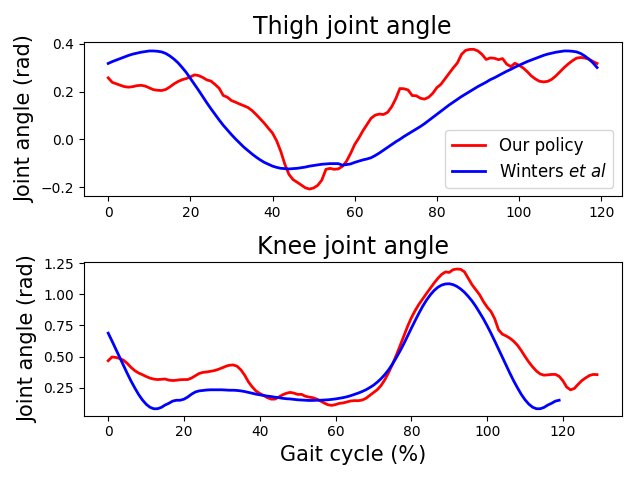}
\caption{Comparison between hip and knee joint angles during walking generated by the policy and human data~\cite{Winter}.}
\label{fig:Jangles}
\end{figure}

\begin{figure}
\centering
\begin{subfigure}{1.0\linewidth}
  \centering
  \includegraphics[width=0.8\linewidth]{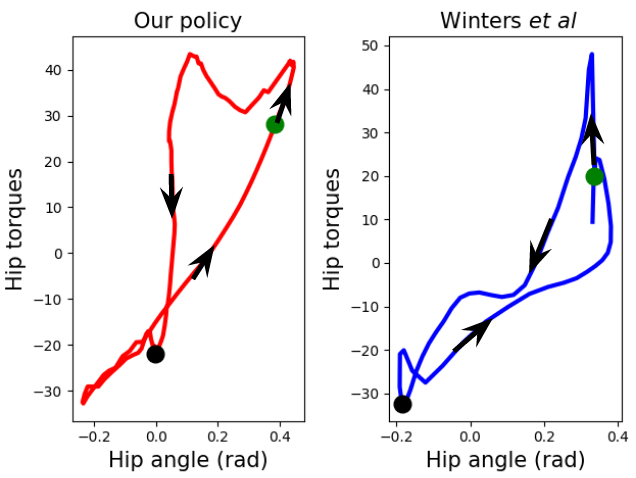}
  \caption{}
  \label{fig:Tauloops}
\end{subfigure}
\begin{subfigure}{1.0\linewidth}
  \centering
  \includegraphics[width=0.8\linewidth]{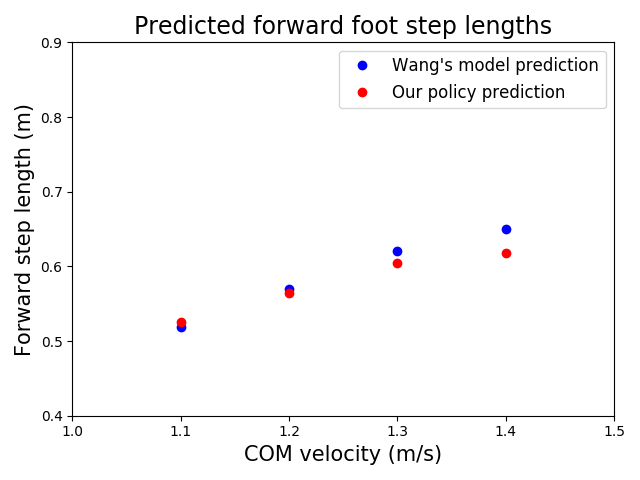}
  \caption{}
  \label{fig:footplacement}
\end{subfigure}
\caption{(a) Comparison of torque loops of a typical trajectory generated by our policy and human data reported by \cite{Winter} at the hip of stance leg during a gait cycle. The green dots indicate the start and the black dots indicate 50\% of the gait cycle. The arrows show the progression of the gait from 0\% to 100\%. (b) Comparison of the forward foot step locations predicted by the policy and by the model reported by Wang \etal \cite{Wang2014} as a function of the COM velocity. }
\label{fig:test}
\end{figure}

We first validate the steady walking behavior of the human policy by comparing it to the data collected from human-subject experiments. \figref{Jangles} shows that the hip and knee joint angles generated by the walking policy well match the data reported in Winter \etal \cite{Winter}. We also compare the ``torque loop'' between the gait generated by our learned policy and the gait recorded from the real world~\cite{Winter}. A torque loop is a plot that shows the relation between the joint degree of freedom and the torque it exerts, frequently used in the biomechanics literature as a metric to quantify human gait. Although the torque loops in \figref{Tauloops} are not identical, both trajectories form loops during a single gait cycle indicating energy being added and removed during the cycle. We also notice that the torque range and the joint angle range are similar. 

In addition, we compare adjusted footstep locations due to external perturbations with the studies reported by Wang \etal \cite{Wang2014}. 
Their findings strongly indicate that the COM dynamics is crucial in predicting the step placement after disturbance that leads to a balanced state. They introduced a model to predict the changes in location of the foot placement of a normal gait as a function of the COM velocity. \figref{footplacement} illustrates the foot placements of our model and the model of Wang \etal against four pushes with different magnitudes in the sagittal plane. For all scenarios, the displacement error is below $4$~cm.


\subsection{Effectiveness of Recovery Policy}

\begin{figure}
\centering
\setlength{\tabcolsep}{3pt}
\begin{tabular}{c c c c}
 
  \includegraphics[width=0.11\textwidth,height=2.1cm]{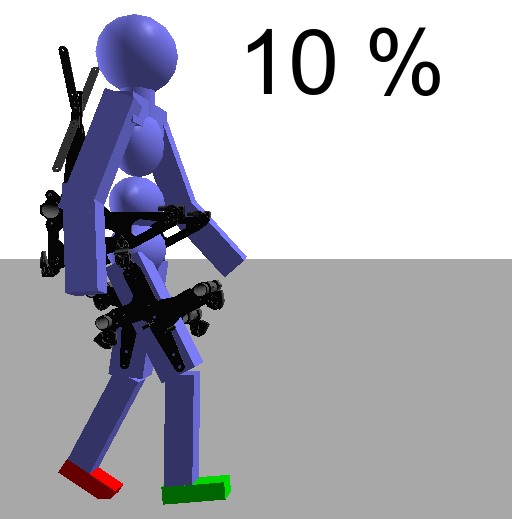}&
  \includegraphics[width=0.11\textwidth,height=2.1cm]{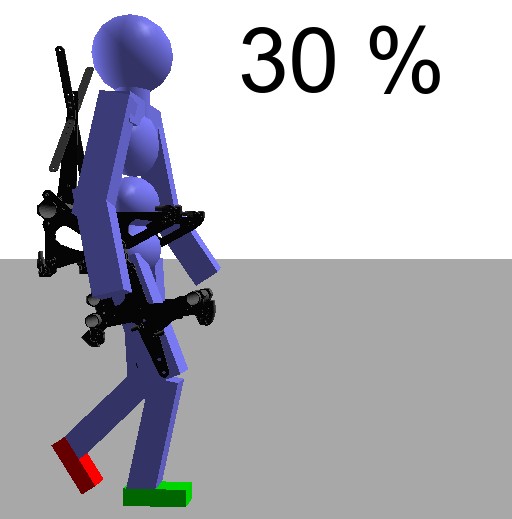}&
  \includegraphics[width=0.11\textwidth,height=2.1cm]{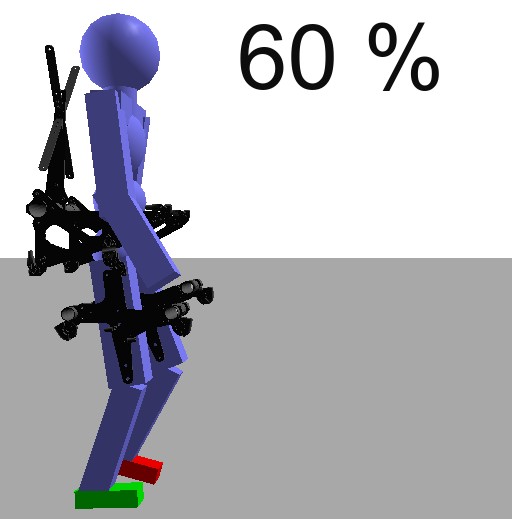}&
  \includegraphics[width=0.11\textwidth,height=2.1cm]{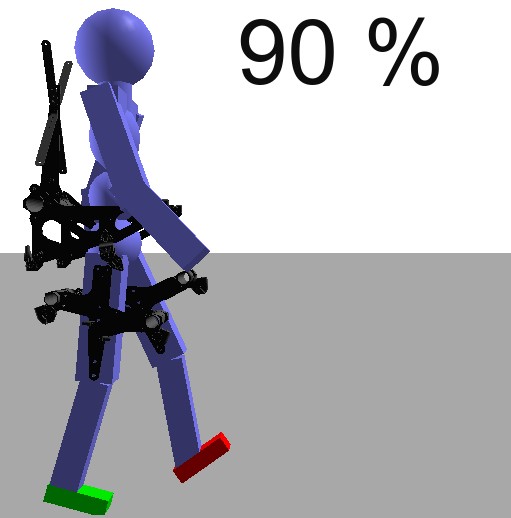} \\ 
  
\end{tabular}
\caption{ Four different timing of the left leg swing phase during which we test the performance of the assistive device. First is at 10\% of the phase and then subsequently 30\%, 60\% and 90\% of the left swing leg.}
\label{fig:phaseDefinition}
\end{figure}

We test the performance of the learned recovery policy in the simulated environment with external pushes. As a performance criterion, we define the \emph{stability region} as a range of external pushes from which the policy can return to the steady gait without falling. For better 2D visualization, we fix the pushes to be parallel to the plane, applied on the same location with the same timing and duration (40 milliseconds). All the experiments in this section use the default sensors and actuators provided by the prototype of the walking device: an IMU, hip joint motor encoders, and hip actuators that control the flexion and extension of the hip.

\begin{figure}
\centering
 \begin{subfigure}{0.8\linewidth}
   \centering
  \includegraphics[width=0.8\linewidth]{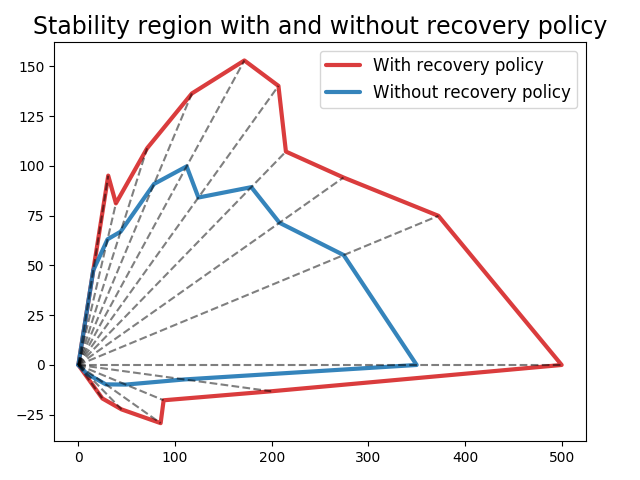}
   \caption{}
 \end{subfigure}

\caption{Stability region with and without the use of a recovery policy. A larger area shows increased robustness to an external push in both magnitude and direction. }
\label{fig:exohelp}
\end{figure}

\begin{figure*}
\centering
\setlength{\tabcolsep}{1pt}
\renewcommand{\arraystretch}{0.7}
\begin{tabular}{c c c c c}
 
  \includegraphics[width=0.19\textwidth,height=2.3cm]{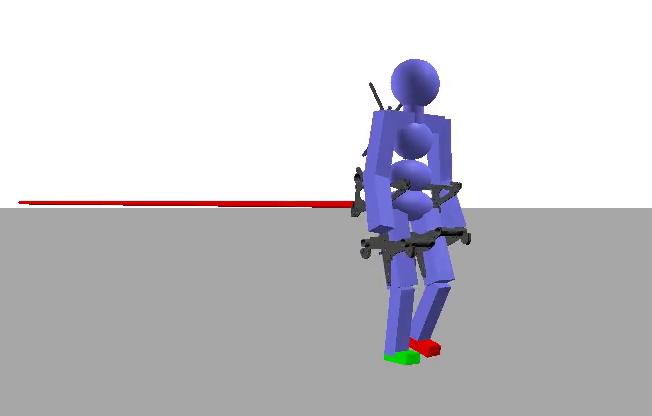}&
  \includegraphics[width=0.19\textwidth,height=2.3cm]
  {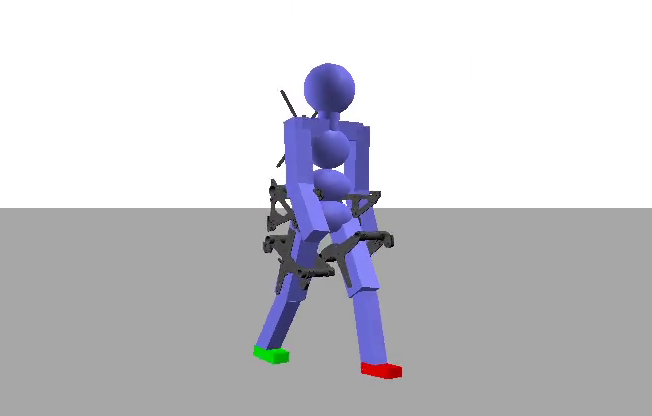}&
  \includegraphics[width=0.19\textwidth,height=2.3cm]{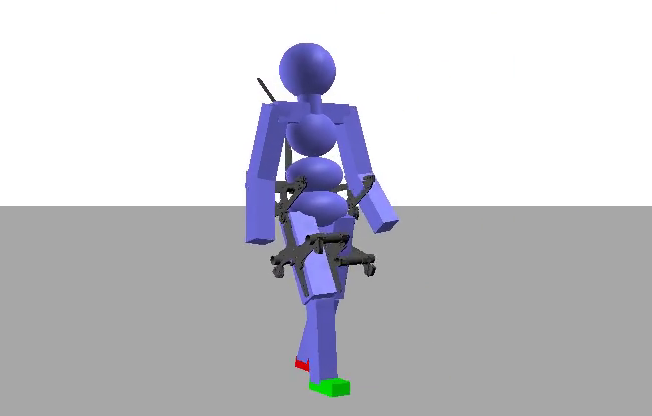}&
  \includegraphics[width=0.19\textwidth,height=2.3cm]{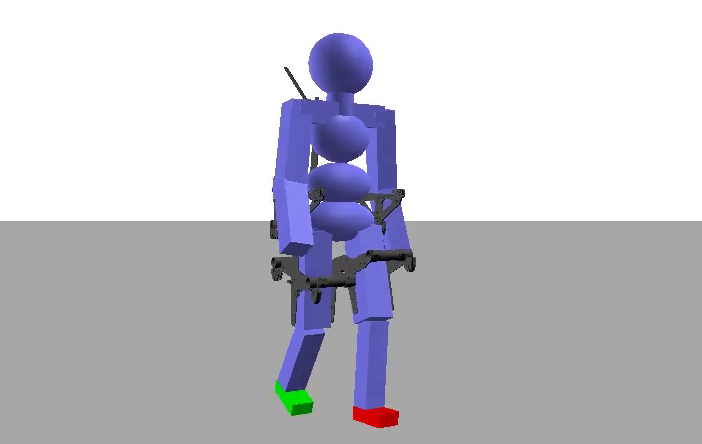}&
  \includegraphics[width=0.19\textwidth,height=2.3cm]{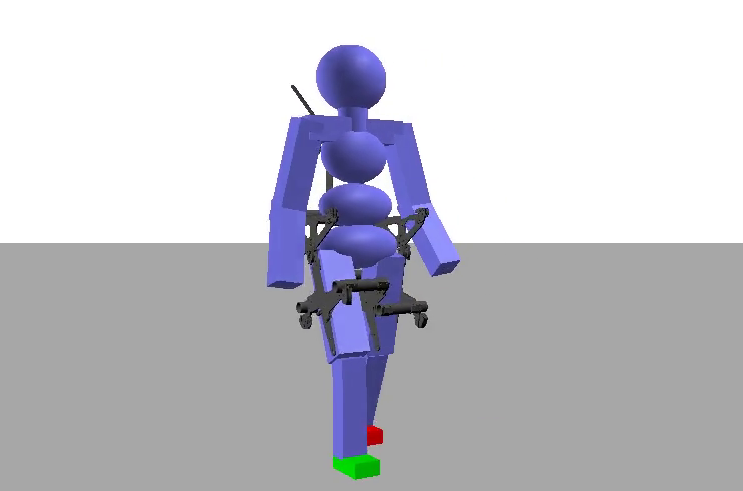} \\ 
  
  \includegraphics[width=0.19\textwidth,height=2.3cm]{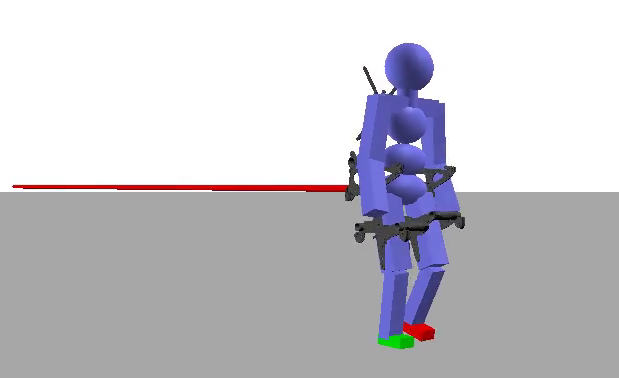}&
  \includegraphics[width=0.19\textwidth,height=2.3cm]
  {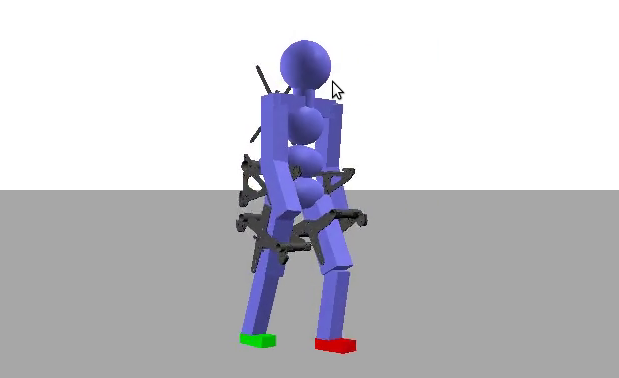}&
  \includegraphics[width=0.19\textwidth,height=2.3cm]{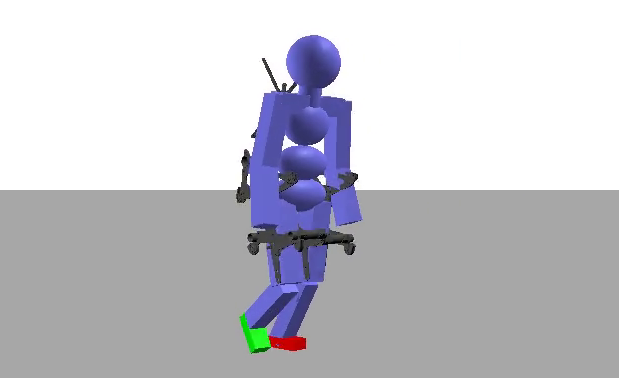}&
  \includegraphics[width=0.19\textwidth,height=2.3cm]{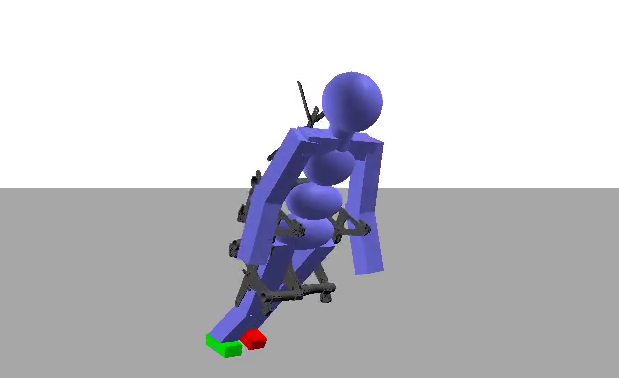}&
  \includegraphics[width=0.19\textwidth,height=2.3cm]{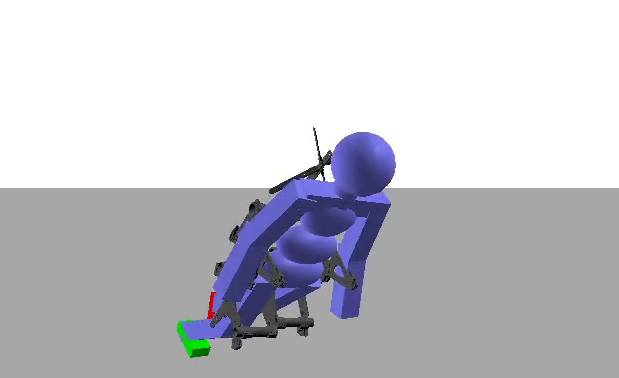} \\
\end{tabular}
\caption{ \textbf{Top:} Successful gait with an assistive device.
\textbf{Bottom:} Unsuccessful gait without an assistive device.}
\label{fig:s}
\end{figure*}

\figref{exohelp} compares the stability region with and without the learned recovery policy. The area of stability region is expanded by 35\% when the recovery policy is used. Note that the stability region has very small coverage on the negative side of y-axis which corresponds to the rightward forces. This is because we push the agent when the swing leg is the left one, making it difficult to counteract the rightward pushes. Figure \ref{fig:s} shows one example of recovery motion.

\begin{figure}
\centering
\includegraphics[width=0.8\linewidth]{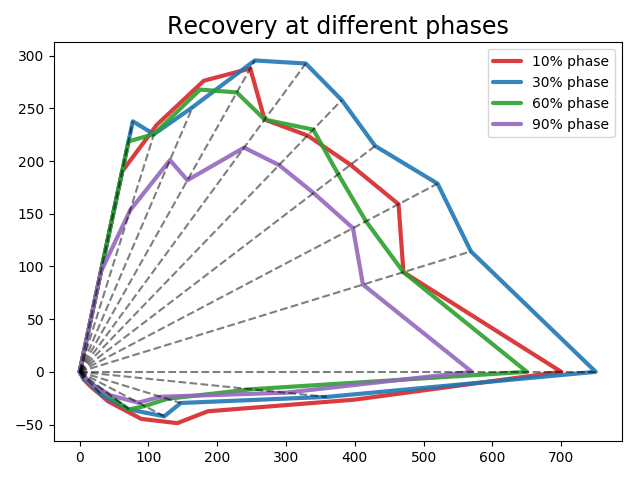}
\includegraphics[width=0.9\linewidth]{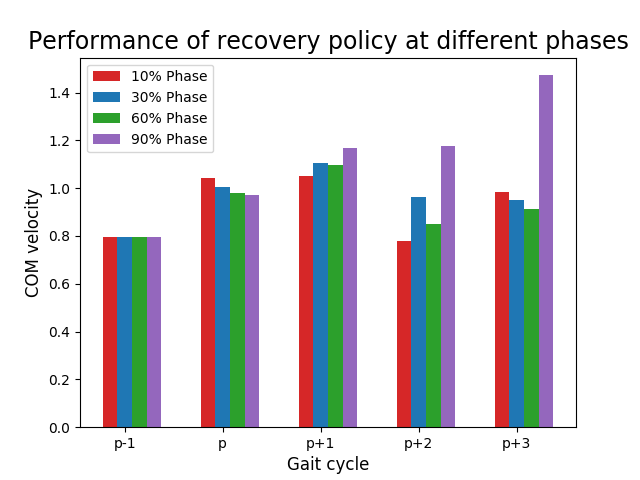}
\caption{Comparison of recovery performance when perturbation is applied at four different phases. \textbf{Top:} Comparison of stability region. \textbf{Bottom:} Comparison of COM velocity across five gait cycles. Perturbation is applied during the gait cycle 'p'. Both plots indicate that our policy is least effective at recovering when the perturbation occurs later in the swing phase.
} 
\label{fig:PhaseEffect}
\end{figure}

The timing of the push in a gait cycle has a great impact on fall prevention. We test our recovery policy with perturbation applied at four different phases during the swing phase (\figref{phaseDefinition}). We found that the stability region is the largest when the push is applied at 30\% of the swing phase and the smallest at 90\% (\figref{PhaseEffect}, Top). This indicates that the perturbation occurs right before heel strike is more difficult to recover than the one occurs in early swing phase possibly due to the lack of time to adjust the foot location. The difference in the stability region is approximately $28$\%. The bottom of \figref{PhaseEffect} shows the impact of the perturbation timing on COM velocity over four gait cycles. The results echo the previous finding as it shows that the agent fails to return to the steady state when the perturbation occurs later in the swing phase.

\begin{figure}
\centering
\includegraphics[width=0.8\linewidth]{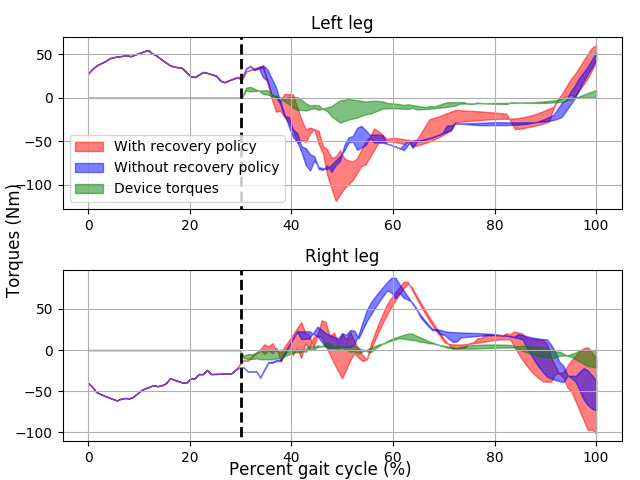}
\vspace{2mm}

\caption{Average torques at the hip joints from $50$ trials with various perturbations. The shaded regions represent the 3-sigma bounds. \textbf{Red:} Joint torques exerted by the human and the recovery policy. \textbf{Blue:} Human joint torques without a recovery policy. \textbf{Green:} Torques produced by a recovery policy.}
\label{fig:Power}
\end{figure}

We also compare the generated torques with and without the recovery policy when perturbation is applied. \figref{Power} shows the torques at the hip joint over the entire gait cycle (not just swing phase). We collect $50$ trajectory for each scenario by applying random forces ranging from $200$N to $800$N at the fixed timing of $30$\% of the gait cycle. The results show that hip torques exerted by the human together with the recovery policy do not change the overall torque profile significantly, suggesting that the recovery policy makes minor modification to the torque profile across the remaining gait cycle, instead of generating a large impulse to stop the fall. We also show that the torque exerted by the recovery policy never exceeds the actuation limits of the device.


\begin{figure}
\centering
\includegraphics[width=0.8\linewidth]{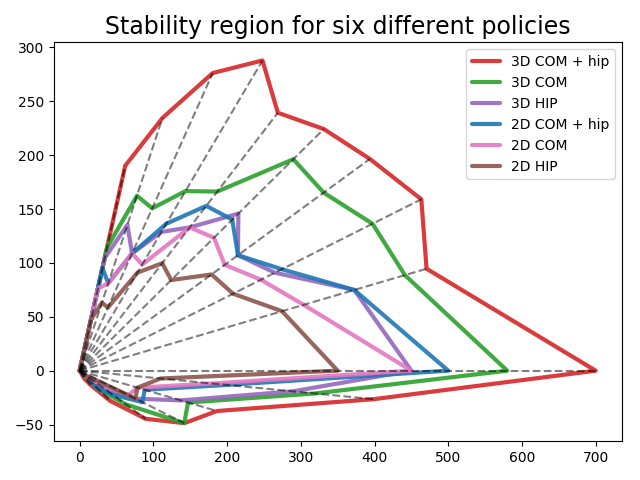}

\caption{Stability region for six policies trained with three sensor configurations and two actuator configurations. }

\label{fig:SensorEffect}
\end{figure}

\subsection{Evaluation of Different Design Choices}




Our method can be used to inform the selection of sensors and actuators when designing a walking device with the capability of fall prevention. We test two versions of actuators: the $2$D hip device can actuate the hip joints only in the sagittal plane while the $3$D device also allows actuation in the frontal plane. We also consider three different configurations of sensors: an inertial measurement unit (IMU) that provides the COM velocity and acceleration, an motor encoder that gives us hip joint angles, and the combination of IMU and motor encoder. In total, we train six different recovery policies with three sensory inputs and two different actuation capabilities. For each sensor configuration, we train a fall predictor using only sensors available to that configuration.

\figref{SensorEffect} shows the stability region for each of the six design configurations. The results indicate that 3D actuation expands the stability region in all directions significantly comparing to 2D actuation, even when the external force lies on the sagittal plane. We also found that the IMU sensor plays a more important role than the motor encoder, which suggests that COM information is more critical than the hip joint angle in informing the action for recovery. The recovery policy performs the best when combining the IMU and the joint encoder, as expected.

%% file: Conclusion.tex
\section{Conclusion and Discussion}

\label{sec:Conc}
We presented an approach to automate the process of augmenting an assistive walking device with ability to prevent falls. Our method has three key components : A human walking policy, fall predictor and a recovery policy. In a simulated environment we showed that an assistive device can indeed help recover balance from a wider range of external perturbations. We introduced \emph{stability region} as a quantitative metric to show the benefit of using a recovery policy. In addition to this, \emph{stability region} can also be used to analyze different design choices for an assistive device. We evaluated six different sensor and actuator configurations. 

In this work, we only evaluated the effectiveness of using a recovery policy for an external push. It would be interesting to extend our work to other kinds of disturbances such as tripping and slipping. Another future direction would like to take is deploying our recovery policy on the real-world assistive device. This would need additional efforts to make sure that our recovery policy also can adjust for the differences in body structure of users.  

\section*{Acknowledgment}

\small

We thank Pawel Golyski and Seung-yong Hyung for their assistance with this work. This work was supported by the Global Research Outreach program of Samsung Advanced Institute of Technology.

%% file: root.bbl
\begin{thebibliography}{10}
\providecommand{\url}[1]{#1}
\csname url@samestyle\endcsname
\providecommand{\newblock}{\relax}
\providecommand{\bibinfo}[2]{#2}
\providecommand{\BIBentrySTDinterwordspacing}{\spaceskip=0pt\relax}
\providecommand{\BIBentryALTinterwordstretchfactor}{4}
\providecommand{\BIBentryALTinterwordspacing}{\spaceskip=\fontdimen2\font plus
\BIBentryALTinterwordstretchfactor\fontdimen3\font minus
  \fontdimen4\font\relax}
\providecommand{\BIBforeignlanguage}[2]{{%
\expandafter\ifx\csname l@#1\endcsname\relax
\typeout{** WARNING: IEEEtran.bst: No hyphenation pattern has been}%
\typeout{** loaded for the language `#1'. Using the pattern for}%
\typeout{** the default language instead.}%
\else
\language=\csname l@#1\endcsname
\fi
#2}}
\providecommand{\BIBdecl}{\relax}
\BIBdecl

\bibitem{rubenstein}
L.~Z. Rubenstein, ``Falls in older people: epidemiology, risk factors and
  strategies for prevention,'' \emph{Age and ageing}, vol.~35, no. suppl\_2,
  pp. ii37--ii41, 2006.

\bibitem{Thatte2019}
N.~S. Nitish~Thatte and H.~Geyer, ``Real-time reactive trip avoidance for
  powered transfemoral prostheses,'' \emph{Robotics Systems and Sciences
  (RSS)}, 2019.

\bibitem{Wu2015}
Q.~Wu, X.~Wang, F.~Du, and X.~Zhang, ``{Design and control of a powered hip
  exoskeleton for walking assistance},'' \emph{International Journal of
  Advanced Robotic Systems}, vol.~12, 2015.

\bibitem{Thatte2015}
\BIBentryALTinterwordspacing
N.~Thatte and H.~Geyer, ``{Toward Balance Recovery with Leg Prostheses using
  Neuromuscular Model Control.}'' \emph{IEEE Transactions on Biomedical
  Engineering}, vol.~PP, no.~99, p.~1, 2015. [Online]. Available:
  \url{http://www.ncbi.nlm.nih.gov/pubmed/26315935}
\BIBentrySTDinterwordspacing

\bibitem{GALLE2017183}
\BIBentryALTinterwordspacing
S.~Galle, W.~Derave, F.~Bossuyt, P.~Calders, P.~Malcolm, and D.~D. Clercq,
  ``Exoskeleton plantarflexion assistance for elderly,'' \emph{Gait and
  Posture}, vol.~52, pp. 183 -- 188, 2017. [Online]. Available:
  \url{http://www.sciencedirect.com/science/article/pii/S0966636216306798}
\BIBentrySTDinterwordspacing

\bibitem{peng2018deepmimic}
X.~B. Peng, P.~Abbeel, S.~Levine, and M.~van~de Panne, ``Deepmimic:
  Example-guided deep reinforcement learning of physics-based character
  skills,'' \emph{ACM Transactions on Graphics (Proc. SIGGRAPH 2018)}, 2018.

\bibitem{YuSIGGRAPH2018}
W.~Yu, G.~Turk, and C.~K. Liu, ``Learning symmetric and low-energy
  locomotion,'' \emph{ACM Transactions on Graphics (Proc. SIGGRAPH 2018)},
  vol.~37, no.~4, 2018.

\bibitem{Winter}
D.~A. Winter, \emph{Biomechanics and motor control of human gait: normal,
  elderly and pathological}.\hskip 1em plus 0.5em minus 0.4em\relax Waterloo
  Biomechanics, 1991.

\bibitem{Wang2014}
Y.~Wang and M.~Srinivasan, ``{Stepping in the direction of the fall: The next
  foot placement can be predicted from current upper body state in steady-state
  walking},'' \emph{Biology Letters}, vol.~10, no.~9, 2014.

\bibitem{YAN2015120}
\BIBentryALTinterwordspacing
T.~Yan, M.~Cempini, C.~M. Oddo, and N.~Vitiello, ``Review of assistive
  strategies in powered lower-limb orthoses and exoskeletons,'' \emph{Robotics
  and Autonomous Systems}, vol.~64, pp. 120 -- 136, 2015. [Online]. Available:
  \url{http://www.sciencedirect.com/science/article/pii/S0921889014002176}
\BIBentrySTDinterwordspacing

\bibitem{Wangpre2015}
S.~{Wang}, L.~{Wang}, C.~{Meijneke}, E.~{van Asseldonk}, T.~{Hoellinger},
  G.~{Cheron}, Y.~{Ivanenko}, V.~{La Scaleia}, F.~{Sylos-Labini},
  M.~{Molinari}, F.~{Tamburella}, I.~{Pisotta}, F.~{Thorsteinsson},
  M.~{Ilzkovitz}, J.~{Gancet}, Y.~{Nevatia}, R.~{Hauffe}, F.~{Zanow}, and
  H.~{van der Kooij}, ``Design and control of the mindwalker exoskeleton,''
  \emph{IEEE Transactions on Neural Systems and Rehabilitation Engineering},
  vol.~23, no.~2, pp. 277--286, March 2015.

\bibitem{JezPre2004}
S.~{Jezernik}, G.~{Colombo}, and M.~{Morari}, ``Automatic gait-pattern
  adaptation algorithms for rehabilitation with a 4-dof robotic orthosis,''
  \emph{IEEE Transactions on Robotics and Automation}, vol.~20, no.~3, pp.
  574--582, June 2004.

\bibitem{BlayaPre}
J.~A. {Blaya} and H.~{Herr}, ``Adaptive control of a variable-impedance
  ankle-foot orthosis to assist drop-foot gait,'' \emph{IEEE Transactions on
  Neural Systems and Rehabilitation Engineering}, vol.~12, no.~1, pp. 24--31,
  March 2004.

\bibitem{WickModel}
A.~{Duschau-Wicke}, J.~{von Zitzewitz}, A.~{Caprez}, L.~{Lunenburger}, and
  R.~{Riener}, ``Path control: A method for patient-cooperative robot-aided
  gait rehabilitation,'' \emph{IEEE Transactions on Neural Systems and
  Rehabilitation Engineering}, vol.~18, no.~1, pp. 38--48, Feb 2010.

\bibitem{KazModel}
\BIBentryALTinterwordspacing
H.~Kazerooni, R.~Steger, and L.~Huang, ``Hybrid control of the berkeley lower
  extremity exoskeleton (bleex),'' \emph{The International Journal of Robotics
  Research}, vol.~25, no. 5-6, pp. 561--573, 2006. [Online]. Available:
  \url{https://doi.org/10.1177/0278364906065505}
\BIBentrySTDinterwordspacing

\bibitem{vanderkooijModel}
H.~{Vallery}, J.~{Veneman}, E.~{van Asseldonk}, R.~{Ekkelenkamp}, M.~{Buss},
  and H.~{van Der Kooij}, ``Compliant actuation of rehabilitation robots,''
  \emph{IEEE Robotics Automation Magazine}, vol.~15, no.~3, pp. 60--69, Sep.
  2008.

\bibitem{Thatte2017}
N.~Thatte, H.~Duan, and H.~Geyer, ``{A Sample-Efficient Black-Box Optimizer to
  Train Policies for Human-in-the-Loop Systems with User Preferences},'' pp.
  1--8, 2017.

\bibitem{lenzi}
T.~{Lenzi}, L.~{Hargrove}, and J.~{Sensinger}, ``Speed-adaptation mechanism:
  Robotic prostheses can actively regulate joint torque,'' \emph{IEEE Robotics
  Automation Magazine}, vol.~21, no.~4, pp. 94--107, Dec 2014.

\bibitem{lenziminjerk}
T.~{Lenzi}, L.~J. {Hargrove}, and J.~W. {Sensinger}, ``Minimum jerk swing
  control allows variable cadence in powered transfemoral prostheses,'' in
  \emph{2014 36th Annual International Conference of the IEEE Engineering in
  Medicine and Biology Society}, Aug 2014, pp. 2492--2495.

\bibitem{Greggvirtual}
R.~D. {Gregg}, T.~{Lenzi}, L.~J. {Hargrove}, and J.~W. {Sensinger}, ``Virtual
  constraint control of a powered prosthetic leg: From simulation to
  experiments with transfemoral amputees,'' \emph{IEEE Transactions on
  Robotics}, vol.~30, no.~6, pp. 1455--1471, Dec 2014.

\bibitem{supimpedence}
F.~{Sup}, H.~{Atakan Varol}, J.~{Mitchell}, T.~J. {Withrow}, and M.~{Goldfarb},
  ``Preliminary evaluations of a self-contained anthropomorphic transfemoral
  prosthesis,'' \emph{IEEE/ASME Transactions on Mechatronics}, vol.~14, no.~6,
  pp. 667--676, Dec 2009.

\bibitem{schulman2015trust}
J.~Schulman, S.~Levine, P.~Abbeel, M.~Jordan, and P.~Moritz, ``Trust region
  policy optimization,'' in \emph{International Conference on Machine
  Learning}, 2015, pp. 1889--1897.

\bibitem{schulman2017proximal}
J.~Schulman, F.~Wolski, P.~Dhariwal, A.~Radford, and O.~Klimov, ``Proximal
  policy optimization algorithms,'' \emph{arXiv preprint arXiv:1707.06347},
  2017.

\bibitem{HAMAYA201767}
\BIBentryALTinterwordspacing
M.~Hamaya, T.~Matsubara, T.~Noda, T.~Teramae, and J.~Morimoto, ``Learning
  assistive strategies for exoskeleton robots from user-robot physical
  interaction,'' \emph{Pattern Recognition Letters}, vol.~99, pp. 67 -- 76,
  2017, user Profiling and Behavior Adaptation for Human-Robot Interaction.
  [Online]. Available:
  \url{http://www.sciencedirect.com/science/article/pii/S0167865517301198}
\BIBentrySTDinterwordspacing

\bibitem{Bingjing2019}
\BIBentryALTinterwordspacing
G.~Bingjing, H.~Jianhai, L.~Xiangpan, and Y.~Lin, ``Human–robot interactive
  control based on reinforcement learning for gait rehabilitation training
  robot,'' \emph{International Journal of Advanced Robotic Systems}, vol.~16,
  no.~2, p. 1729881419839584, 2019. [Online]. Available:
  \url{https://doi.org/10.1177/1729881419839584}
\BIBentrySTDinterwordspacing

\bibitem{Shiratori2009}
\BIBentryALTinterwordspacing
T.~Shiratori, B.~Coley, R.~Cham, and J.~K. Hodgins, ``{Simulating Balance
  Recovery Responses to Trips based on Biomechanical Principles},'' \emph{Proc.
  of ACM SIGGRAPH/Eurographics Symposium on Computer Animation 2009}, pp.
  37--46, 2009. [Online]. Available:
  \url{http://graphics.cs.cmu.edu/projects/trip/}
\BIBentrySTDinterwordspacing

\bibitem{Lee:2019:SMH:3306346.3322972}
\BIBentryALTinterwordspacing
S.~Lee, M.~Park, K.~Lee, and J.~Lee, ``Scalable muscle-actuated human
  simulation and control,'' \emph{ACM Trans. Graph.}, vol.~38, no.~4, pp.
  73:1--73:13, Jul. 2019. [Online]. Available:
  \url{http://doi.acm.org/10.1145/3306346.3322972}
\BIBentrySTDinterwordspacing

\bibitem{Jiang:2019:SBR:3306346.3322966}
\BIBentryALTinterwordspacing
Y.~Jiang, T.~Van~Wouwe, F.~De~Groote, and C.~K. Liu, ``Synthesis of
  biologically realistic human motion using joint torque actuation,'' \emph{ACM
  Trans. Graph.}, vol.~38, no.~4, pp. 72:1--72:12, Jul. 2019. [Online].
  Available: \url{http://doi.acm.org/10.1145/3306346.3322966}
\BIBentrySTDinterwordspacing

\bibitem{Lockhart2012}
T.~E. Lockhart, ``{Biomechanics of Human Gait - Slip and Fall Analysis},''
  \emph{Encyclopedia of Forensic Sciences: Second Edition}, vol.~2, pp.
  466--476, 2012.

\bibitem{Moyer2009}
B.~Moyer, M.~Redfern, and R.~Cham, ``Biomechanics of trailing leg response to
  slipping-evidence of interlimb and intralimb coordination,'' \emph{Gait and
  posture}, vol.~29, no.~4, pp. 565--570, 2009.

\bibitem{connor09}
K.~A. O'Connor~SM, ``Direction-dependent control of balance during walking and
  standing,'' \emph{J Neurophysiol. 2009;102(3):1411–1419.
  doi:10.1152/jn.00131.2009}, 2009.

\bibitem{Hof}
\BIBentryALTinterwordspacing
A.~L. Hof, S.~M. Vermerris, and W.~A. Gjaltema, ``Balance responses to lateral
  perturbations in human treadmill walking,'' \emph{Journal of Experimental
  Biology}, vol. 213, no.~15, pp. 2655--2664, 2010. [Online]. Available:
  \url{https://jeb.biologists.org/content/213/15/2655}
\BIBentrySTDinterwordspacing

\bibitem{lee2018dart}
J.~Lee, M.~X. Grey, S.~Ha, T.~Kunz, S.~Jain, Y.~Ye, S.~S. Srinivasa,
  M.~Stilman, and C.~K. Liu, ``Dart: Dynamic animation and robotics toolkit,''
  \emph{The Journal of Open Source Software}, vol.~3, no.~22, p. 500, 2018.

\bibitem{caputo}
\BIBentryALTinterwordspacing
J.~M. Caputo and S.~H. Collins, ``{A Universal Ankle–Foot Prosthesis Emulator
  for Human Locomotion Experiments},'' \emph{Journal of Biomechanical
  Engineering}, vol. 136, no.~3, 02 2014, 035002. [Online]. Available:
  \url{https://doi.org/10.1115/1.4026225}
\BIBentrySTDinterwordspacing

\bibitem{ExoDesign}
K.~A. Witte and S.~H. Collins, ``Design of lower-limb exoskeletons and emulator
  systems.'' in \emph{Wearable Robotics}, R.~J. Ferguson, P.~W., Ed.\hskip 1em
  plus 0.5em minus 0.4em\relax Elsevier, 2019, ch.~22.

\bibitem{witte}
K.~A. {Witte}, J.~{Zhang}, R.~W. {Jackson}, and S.~H. {Collins}, ``Design of
  two lightweight, high-bandwidth torque-controlled ankle exoskeletons,'' in
  \emph{2015 IEEE International Conference on Robotics and Automation (ICRA)},
  May 2015, pp. 1223--1228.

\end{thebibliography}
